\pdfoutput=1

\documentclass[11pt]{article}

\usepackage[preprint]{acl}

\usepackage{times}
\usepackage{url}
\usepackage{latexsym}

\usepackage[T1]{fontenc}

\usepackage[utf8]{inputenc}

\usepackage{microtype}

\usepackage{inconsolata}
\usepackage{graphicx}
\usepackage{booktabs,tabularx,caption}
\usepackage{multirow}
\usepackage{amsfonts} 
\usepackage{amsmath}
\usepackage{bbding}
\usepackage{enumitem}
\setitemize[1]{itemsep=2pt,partopsep=0pt,parsep=\parskip,topsep=2pt}

\title{KB-Plugin: A Plug-and-play Framework for Large Language Models to \\Induce Programs over Low-resourced Knowledge Bases}

\author{
 Jiajie~Zhang$^{1}$\hspace{0.3em}, Shulin~Cao$^{1}$, Linmei~Hu$^{2}$, Ling~Feng$^1$, Lei~Hou$^1$\thanks{\ \ Corresponding author.}, Juanzi~Li$^1$\\ 
 $^1$Tsinghua University \quad 
 $^2$Beijing Institute of Technology \\
\texttt{zhangjj23@mails.tsinghua.edu.cn}\\
}

\begin{document}
\maketitle
\begin{abstract}

Program induction (PI) has become a promising paradigm for using knowledge bases (KBs) to help large language models (LLMs) answer complex knowledge-intensive questions. 
Nonetheless, PI typically relies on a large number of parallel question-program pairs to make the LLM aware of the schema of the given KB, and is thus challenging for many low-resourced KBs that lack annotated data. 
To this end, we propose \textbf{KB-Plugin}, a plug-and-play framework that enables LLMs to induce programs over any low-resourced KB. 
Firstly, KB-Plugin adopts self-supervised learning to encode the detailed schema information of a given KB into a pluggable module, namely \textbf{schema plugin}. Secondly, KB-Plugin utilizes abundant annotated data from a rich-resourced KB to train another pluggable module, namely \textbf{PI plugin}, which can help the LLM extract question-relevant schema information from the schema plugin of any KB and utilize this information to induce programs over this KB. 
Experiments on five heterogeneous KBQA datasets show that KB-Plugin achieves better or comparable performance with 25$\times$ smaller backbone LLM compared to SoTA PI methods for low-resourced KBs, and even approaches the performance of supervised methods. Our code and data are available at \url{https://github.com/THU-KEG/KB-Plugin}.

\end{abstract}

\begin{figure}[t]
    \centering
    \includegraphics[width=0.49\textwidth]{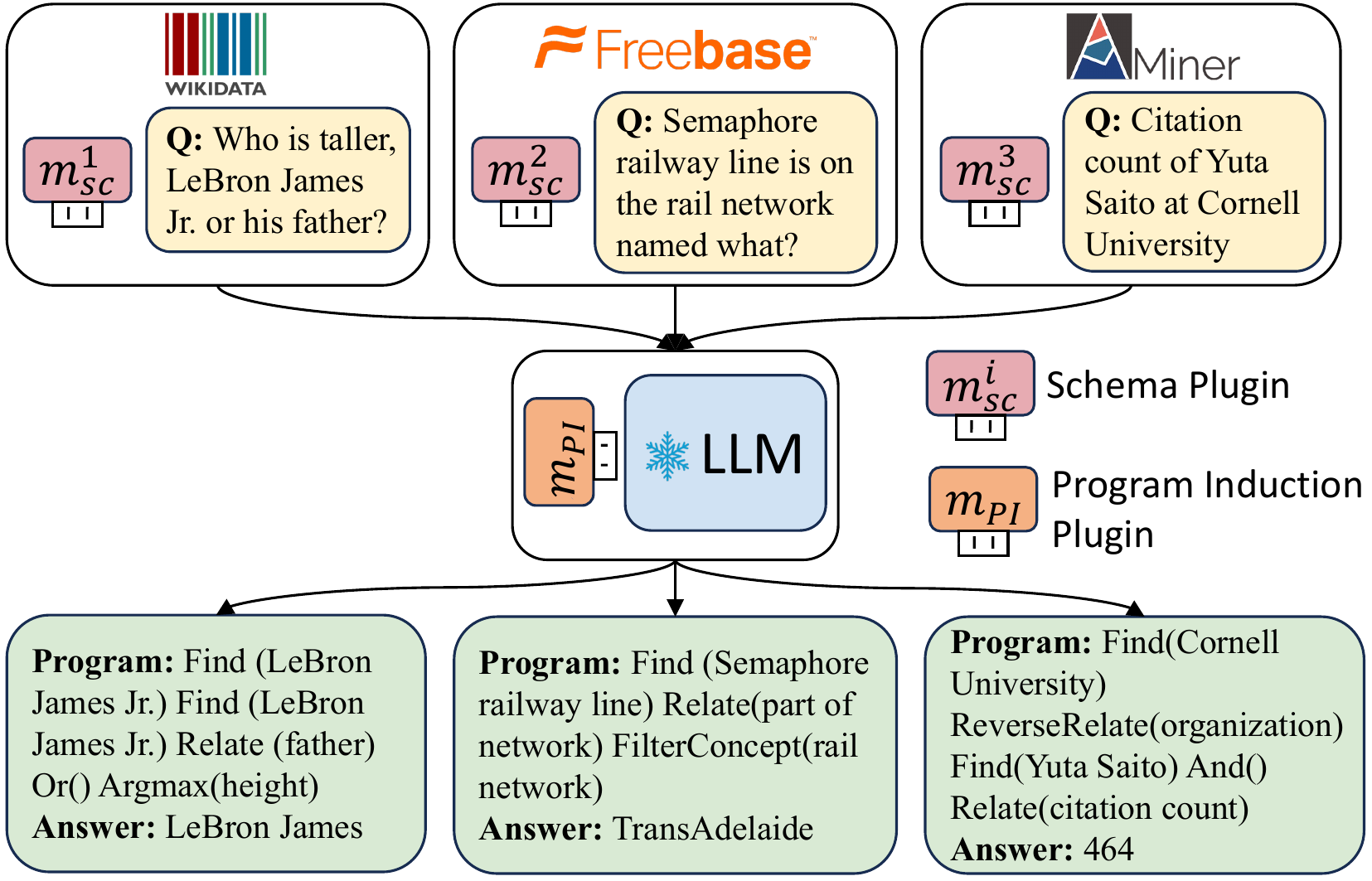}
    \caption{Illustration of KB-Plugin. By simply plugging the schema plugin of a KB and the PI plugin, the LLM is injected with the schema information of this KB and the ability to induce programs over it. }
    \label{fig_first}
\end{figure}

\section{Introduction}
\label{sec_intro}
Recently, the usage of knowledge bases (KBs) as external resources to assist large language models (LLMs)~\cite{gpt3, zhao2023} in answering complex knowledge-intensive questions has gained increasing study~\cite{pan2023, cok, structgpt}. Among various methods, program induction (PI) has emerged as a promising paradigm due to its good interpretability and capacity to support complex reasoning operations~\cite{kqapro, pangu, cok}. Given a KB, PI methods employ LLMs to convert a question into a multi-step program (e.g., KoPL~\cite{kqapro} and S-expression~\cite{graphq}), whose execution against the KB produces the answer. Despite strong capacity, most PI methods rely on individual training for each KB using a large number of manually annotated question-program pairs~\cite{unifiedskg, cok, chatkbqa}. As for many low-resourced KBs that lack program annotations, how to enable LLMs to utilize their knowledge via PI remains a challenging problem.

Recent studies~\cite{program_transfer, kb-binder} have indicated that the mapping from questions to program sketches (i.e., composed functions without arguments, such as \texttt{Find}$ \rightarrow$ \texttt{Relate}$ \rightarrow$ \texttt{FilterConcept}) primarily correlates with language compositional structures and is thus transferable across KBs. Hence the main challenge for PI over low-resourced KBs is to determine the argument for each function~\cite{arcaneqa}, which requires LLMs to link natural language in a question to corresponding schema items (i.e., pre-defined relations and concepts) in the KB (e.g., in Fig~\ref{fig_first}, the relation \textit{``part of network''} and the concept \textit{``rail network''} are arguments of function \texttt{Relate} and \texttt{FilterConcept}, respectively), so it is important to provide LLMs adequate information of each schema item. A straightforward approach is to directly feed all the schema information to the LLM via a prompt. However, the broad schema of KBs and limited context windows of LLMs make this infeasible~\cite{kb-binder}.

Regarding the above challenges, we are inspired by recent studies that claim that the parameters of LLMs can encode task-specific knowledge~\cite{kgt5, SKILL, skill_neuron}.  
Our basic idea is to encode schema information of a KB into the parameters of a pluggable module (e.g., LoRA~\cite{lora}), namely \textbf{schema plugin}, and use another pluggable module, namely \textbf{PI plugin}, to help the LLM capture question-relevant schema information from the schema plugin and utilize this information to induce programs. As illustrated in Fig.~\ref{fig_first}, by simply plugging the schema plugin of a KB and the PI plugin, the LLM is injected with the schema information of this KB and the ability to induce programs over it. We name this framework~\textbf{KB-Plugin}. To implement KB-Plugin, there remain two key problems: (1) By what task can sufficient information about each schema item in a KB be encoded into its schema plugin? (2) Without annotated data from the low-resource KBs, how can the PI plugin learn to extract and utilize question-relevant schema information from their schema plugins to induce programs over these KBs?

To address the above problems, we propose a novel plugin learning and transfer framework. First, inspired by prior studies~\cite{transe, transr} which show that schema items in a KB can be well represented by fact triples involving them, we propose to learn schema plugins via a self-supervised triple completion task. Specifically, given a KB, we plug a schema plugin into the LLM and tune the plugin to enable the LLM to complete relevant triples for each schema item in the KB. In this way, the detailed schema information can be encoded into this schema plugin. As for PI plugin learning, inspired by ~\citet{program_transfer}, we utilize abundant program annotations from a rich-resourced KB. Specifically, we use this KB to generate multiple KBs with different schemas via alias replacement and train a schema plugin for each of them. Given a training question, we plug these schema plugins alone with the PI plugin into the LLM in turn and train the PI plugin to make the LLM generate the correct program whose arguments conform to the currently plugged schema plugin. In this way, the PI plugin is forced to learn the skills of extracting and utilizing question-relevant schema information from the plugged schema plugin for PI over the corresponding KB. 
Besides, since the PI plugin is trained to be compatible with different schema plugins, it can be directly transferred to other low-resourced KBs and generalize well with their schema plugins, even if most schema items in these KBs are unseen during its training.

In experiments, we take Wikidata-based KQA Pro as the rich-resourced KB to train the PI plugin and evaluate our framework on three Freebase-based datasets (WebQSP, GraphQ, and GrailQA) and two domain-specific datasets (MetaQA for movie domain and SoAyBench for academic domain). The results show that KB-Plugin achieves better or comparable performance with 25$\times$ smaller backbone LLM compared to SoTA PI methods for low-resource KBs. On GraphQ, GrailQA, and MetaQA, KB-Plugin even surpasses the performance of several supervised methods. 

\textbf{Our contributions} include: (1) proposing KB-Plugin, a novel plug-and-play framework that enables LLMs to induce programs over any low-resourced KB; (2) empirical validation of the efficacy of KB-Plugin through comprehensive experiments on five heterogeneous KBQA datasets.

\section{Related Work}

\noindent\textbf{Low-resourced Program Induction. }
\label{sec_low_PI}
Recently, there have emerged three types of PI methods for low-resourced KBs that lack program annotations, but each of them has limitations: (1) Few-shot program generation methods~\cite{pangu, kb-binder} utilize in-context learning ability of LLMs to induce programs with a handful of demonstrations. However, they can only determine function arguments based on the schema item names due to limited context windows, so they face challenges in distinguishing similar schema items. They also suffer from long inference time due to excessive LLM calls or executing a vast number of potential programs; 
(2) Few-shot data generation methods~\cite{flexkbqa} also employ in-context learning with LLMs to convert automatically sampled programs into questions, and train a smaller PI model using the generated question-program pairs. Nonetheless, the generated questions may not align with programs and often lack diversity due to the limited number of program templates; (3) Similar to us, program transfer methods~\cite{program_transfer} also leverage program annotations from a rich-resourced KB to aid PI for low-resourced KBs. However, they mainly focus on program sketch transfer and perform poorly without fine-tuning using annotated question-answer pairs from low-resourced KBs to adapt to their schemas. While KB-plugin obviates the reliance on any annotated data from low-resourced KBs, thereby enabling LLMs to easily utilize their knowledge.

\noindent\textbf{Plug-and-Play Modules for LLMs. }
In recent years, various parameter-efficient modules have been proposed to adapt LLMs to different downstream tasks~\cite{lester21, lora, prefix-tuning, adapterfusion} . These modules show plug-and-play characteristics and can inject task-specific knowledge and skills into LLMs~\cite{doc-plugin, zhang2023}. Some researchers also found that pluggable modules for similar tasks encode knowledge and skills into the parametric space in similar ways~\cite{intrinsic-dimension, prompt_transfer}, providing basic rationality for the transferability of our PI plugin. 
\section{Problem Formulation}
In this section, we first provide some necessary definitions and then formulate our task.

\noindent \textbf{Knowledge Base}. A knowledge base (KB) can be formalized as $\mathcal{KB} = \{\mathcal{C}, \mathcal{E}, \mathcal{R}, \mathcal{T}\}$, where $\mathcal{C}$,  $\mathcal{E}$, $\mathcal{R}$ and $\mathcal{T}$ represent the sets of concepts, entities, relations and fact triples, respectively.
Specifically, $\mathcal{R} = \{r_e, r_c\} \cup \mathcal{R}_l$, where $r_e$ is \textit{``instance of''}, $r_c$ is \textit{``subclass of''}, and $\mathcal{R}_l$ is the set of other general relations. Correspondingly, $\mathcal{T}$ can be divided into there disjoint subsets: (1) \textit{``instance of''} triples $\mathcal{T}_e = \{(e, r_e, c) | e \in \mathcal{E}, c \in \mathcal{C}\}$; (2) \textit{``subclass of''} triples $\mathcal{T}_c = \{(c_{i}, r_c, c_{j}) | c_{i}, c_{j} \in \mathcal{C}\}$; (3) relational triples $\mathcal{T}_l = \{(e_i, r, e_j) | e_i, e_j \in \mathcal{E}, r \in \mathcal{R}_{l}\}$.  
Elements in $\mathcal{C}$ and $\mathcal{R}$ are also called the schema items of $\mathcal{KB}$.

\noindent \textbf{Program Induction}. Given a KB $\mathcal{KB}$ and a natural language question $x = \left\langle w_1, w_2, \cdots, w_{|x|} \right\rangle$, program induction (PI) aims to convert $x$ into a program $y$, which would return the correct answer when executed against $\mathcal{KB}$. Formally, $y$ is composed of functions that take a specific type of arguments, and can be serialized as $y=\left\langle f_1(arg_1), \cdots, f_t(arg_t), \cdots, f_{|y|}(arg_{|y|}) \right\rangle, f_t \in \mathcal{F}, arg_t \in \mathcal{E}\cup\mathcal{C}\cup\mathcal{R}\cup\{\emptyset\}$. Here, $\mathcal{F}$ is a set of pre-defined functions that cover basic reasoning operations on KBs. In this work, we use KoPL~\cite{kqapro} as our programming language.

\noindent \textbf{Task Formulation}. Suppose we have access to (1) source KB $\mathcal{KB}^{S}$ and source domain data $\mathcal{D}^{S}=\{(x_i^{S}, y_i^{S})\}_{i=1}^{n^{S}}$, which are question-program pairs for $\mathcal{KB}^{S}$; (2) target KB $\mathcal{KB}^{T}$, which is low-resourced and has no annotated data. The goal is to learn a PI model $M_{PI}^T$ that can translate a question $x^T$ for $\mathcal{KB}^{T}$ into program $y^T$, whose execution on $\mathcal{KB}^{T}$ produces the correct answer. 
\section{Methodology}

\begin{figure*}
    \centering
    \includegraphics[width=\textwidth]{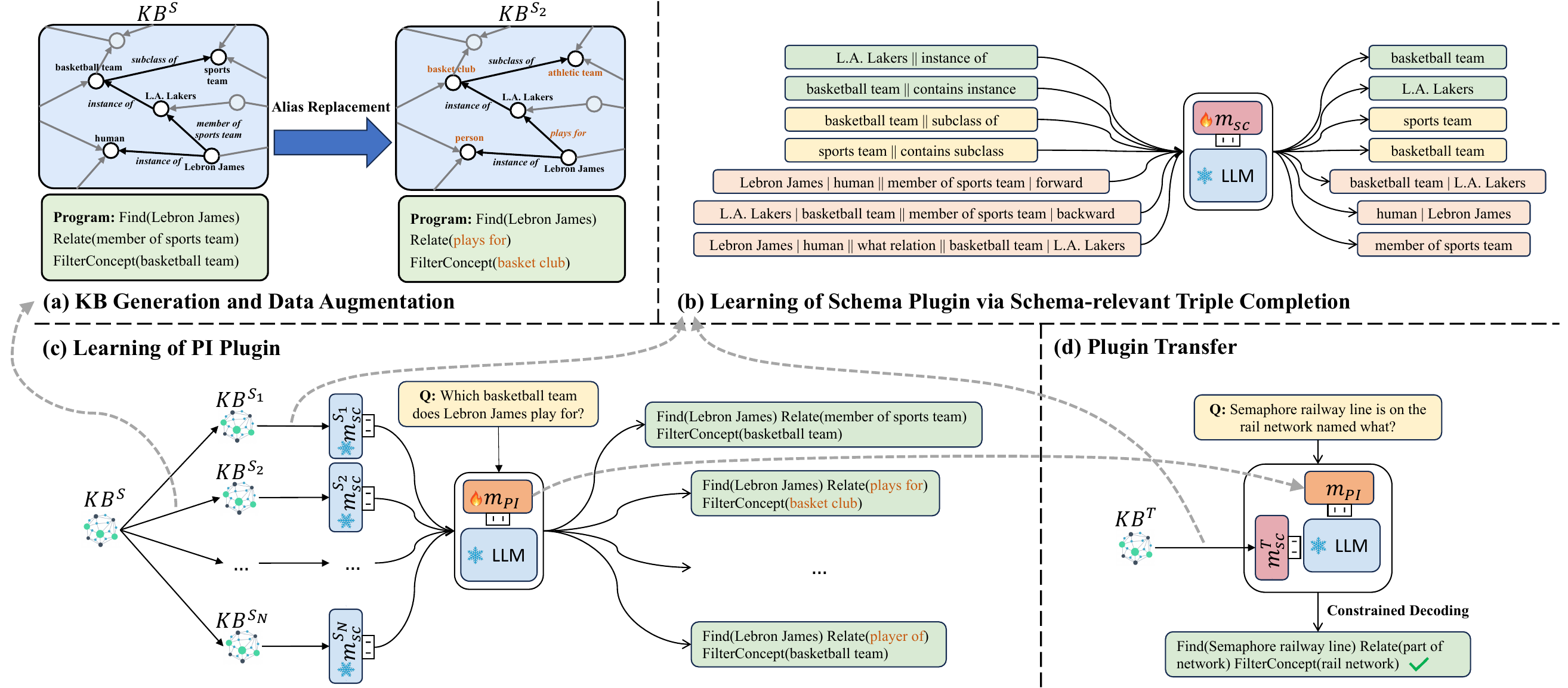}
    \caption{Overview of our plugin learning and transfer framework: (a) Generate multiple source KBs with different schemas and augmented source domain data via alias replacement; (b) Learn an individual schema plugin for each source KB and the target KB via self-supervised schema-relevant triple completion task; (c) Train the PI plugin by inducing program for each source KB when plugging it into the LLM along with the corresponding schema plugin. (d) Transfer the PI plugin by plugging it into the LLM with the schema plugin of the target KB and inducing programs over the target KB with constrained decoding.
    }
    \label{fig_framework}
\end{figure*}

As mentioned in the introduction, to enable a LLM $M$ to induce programs over low-resourced $\mathcal{KB}^T$, KB-Plugin learns two types of pluggable modules for $M$: 
(1) KB-specific \textbf{schema plugin} $m_{sc}$, which stores information of schema items of a given KB within its parameters; 
(2) KB-transferable \textbf{PI plugin} $m_{PI}$, which encodes the skill of inducing programs over any KB by extracting and utilizing question-relevant schema information from the schema plugin of this KB. It is trained with $\mathcal{KB}^S$ and $\mathcal{D^S}$ but can be directly transferred to $\mathcal{KB}^T$. 
The final PI model for $\mathcal{KB}^T$ can be formulated as 
\begin{equation}
    M_{PI}^T=\mathrm{plug}(M, \{m_{sc}^T, m_{PI}\}),
\end{equation}
where $m_{sc}^T$ is the schema plugin of $\mathcal{KB}^T$ and $\mathrm{plug}(M, \{\cdot\})$ means plugging the plugins in $\{\cdot\}$ into $M$. 

In the following, we will first introduce the architecture of two types of plugins, then present our plugin learning and transfer framework.

\subsection{Plugin Architecture}
A host of studies have demonstrated that knowledge and skills can be encapsulated within the parameters of LLMs~\cite{kgt5, SKILL, skill_neuron}. 
Inspired by this, we implement both schema plugin and PI plugin with LoRA~\cite{lora}, a popular type of pluggable module for LLMs with a few trainable parameters. 

Specifically, let $L_M$ be the set of weight matrices in the self-attention modules and MLP modules of a LLM $M$. For each $W_i\in \mathbb{R}^{d\times k}$ in $L_M$, LoRA modifies its forward pass from $h = W_i x$ to $h = (W_i+A_iB_i) x$, where $A_i \in \mathbb{R}^{d\times r}$ and $B_i \in \mathbb{R}^{r\times k}$ are two matrices with rank $r\ll\min(d, k)$. A LoRA plugin $m_j$ is thus defined as 
\begin{equation}
    m_j = \{(A_i^{m_j}, B_i^{m_j}) | W_i \in L_M\},
\end{equation}
and $\mathrm{plug}(M, \{m_1, \dots, m_N\})$ means replacing all $W_i \in L_M$ with $W_i + \sum_{j=1}^N A_i^{m_j} B_i^{m_j}$. 
If we train $M'=\mathrm{plug}(\mathrm{fz}(M), \{\mathrm{fz}(m_1), \dots, \mathrm{fz}(m_{N-1}), m_N\})$ on a certain task, where $\mathrm{fz}(\cdot)$ represents parameter freezing, knowledge and skills related to this task will be encoded within $m_N$. Although other parameter-efficient pluggable modules such as prefix-tuning~\cite{prefix-tuning} can also serve as our plugin modules, the advantages of LoRA are that it does not increase input length or inference latency.

\subsection{Plugin Learning and Transfer Framework}
There are two primary challenges for learning schema plugins and the PI plugin: (1) How to encode sufficient information about each schema item of a KB into a schema plugin? (2) How to ensure that the PI plugin can extract and utilize useful schema information for program induction from schema plugins of different KBs, instead of ignoring the schema plugin entirely, directly learning to induce program over source KB during training, and consequently losing transferability? 

To handle these challenges, we propose a novel plugin learning and transfer framework, which is illustrated in Fig.~\ref{fig_framework} and contains four steps: (1) Generate multiple source KBs $\mathcal{KB}^{S_1}, \dots, \mathcal{KB}^{S_N}$ with different schemas and augmented data $\mathcal{D}^S_a = \{(x_j^S, y_j^{S_1}, \dots, y_j^{S_N})\}_{j=1}^{n^S}$ based on $\mathcal{KB}^S$ and $D^S$ via alias replacement, where $y_j^{S_i}$ is the golden program for question $x_j^S$ on $\mathcal{KB}^{S_i}$; (2) Learn individual schema plugin  $m_{sc}^{S_i}$ for each $\mathcal{KB}^{S_i}$ via self-supervised schema-relevant triple-completion task; (3) Train PI plugin $m_{PI}$ by requiring $M_{PI}^{S_1}, \dots, M_{PI}^{S_N}$ to generate $y_j^{S_1}, \dots, y_j^{S_N}$ given $x_j^S$, respectively, where $M_{PI}^{S_i}=\mathrm{Plug}(\mathrm{fz}({M}), \{\mathrm{fz}(m_{sc}^{S_i}), m_{PI}\})$, so that $m_{PI}$ is forced to extract and utilize schema information from each $m_{sc}^{S_i}$; (4) Learn schema plugin $m_{sc}^{T}$ for $\mathcal{KB}^T$ using the same method in (2) and take $M_{PI}^T=\mathrm{plug}(M, \{m_{sc}^T, m_{PI}\})$ as the final PI model for $\mathcal{KB}^T$. We will introduce each step in detail in the following. 

\subsubsection{KB Generation and Data Augmentation}
We utilize the aliases of each schema item to generate multiple KBs with different schemas based on $\mathcal{KB}^{S}=\{\mathcal{C}^S, \mathcal{E}^S, \mathcal{R}^S, \mathcal{T}^S\}$. As shown in Fig.~\ref{fig_framework}(a), for each schema item $v\in \mathcal{C}^S \cup \mathcal{R}^S$, we replace $v$ with $v_i$, a randomly chosen alias of $v$, and record $a_i(v)=v_i$. For example, the concept ``\textit{basketball team}'' can be replaced with ``\textit{basket club}'' and the relation ``\textit{member of sports team}'' can be replaced with ``\textit{plays for}''. Relevant triples in $\mathcal{T}^S$ are also modified with the same alias. In this way, $\mathcal{KB}^{S_i}$ that has a different schema than $\mathcal{KB}^S$ is created. In practice, we let $\mathcal{KB}^{S_1}=\mathcal{KB}^{S}$ and repeat above process $N-1$ times to generate $\mathcal{KB}^{S_2}, \dots, \mathcal{KB}^{S_N}$. 

Similarly, for each question-program pair $(x^S_j, y^S_j)\in \mathcal{D}^S$, suppose $y^S_j = \left\langle f_1(arg_1), \cdots, f_t(arg_t), \cdots, f_{|y_j^S|}(arg_{|y_j^S|}) \right\rangle$, we replace every $arg_t\in \mathcal{C}^S \cup \mathcal{R}^S$ with $a_i(arg_t)$ to obtain $y^{S_i}_j$, which is the correct program for $x_j^S$ executable on $\mathcal{KB}^{S_i}$. We repeat the process for $\mathcal{KB}^{S_1}, \dots, \mathcal{KB}^{S_N}$ to obtain augmented data $\mathcal{D}^S_a = \{(x_j^S, y_j^{S_1}, \dots, y_j^{S_N})\}_{j=1}^{n^S}$.

\subsubsection{Learning of Schema Plugin}
\label{sec_schema_plugin}
Many studies about knowledge graph embedding show that the information of schema items in a KB can be represented by not only their names but also triples containing them~\cite{transe, transc}. Inspired by this, we propose to encode schema information into schema plugins via a self-supervised triple completion task. As illustrated in Fig.~\ref{fig_framework}(b), 
to learn the schema plugin $m_{sc}$ for a given KB $\mathcal{KB}=\{\mathcal{C}, \mathcal{E}, \mathcal{R}, \mathcal{T}\}$, where $\mathcal{T}=\mathcal{T}_e\cup\mathcal{T}_c\cup\mathcal{T}_l$, 
we train $M_{sc}=\mathrm{Plug}(\mathrm{fz}(M), m_{sc})$ to complete relevant triples for each concept and relation in $\mathcal{KB}$ in sequence-to-sequence form as follows.

First, for each concept $c\in \mathcal{C}$, we require $M_{sc}$ to complete relevant \textit{``instance of''} triples to aggregate the semantic features of entities belonging to $c$. Specifically, we sample $K$ triples $(e_k, \textit{instance of}, c)$ from $\mathcal{T}_e$ (see Appendix~\ref{app_sampling_strategy} for detailed sampling strategy), and use each sampled triple to construct two pairs of verbalized queries and answer as the inputs and expected outputs for $M_{sc}$: 
\begin{itemize}
    \item ``$\langle e_k\rangle$ || \textit{instance of}'' $\rightarrow$ ``$\langle c\rangle$'';
    \item ``$\langle c\rangle$ || \textit{contains instance}'' $\rightarrow$ ``$\langle e_k\rangle$''.
\end{itemize}
Here, $\langle e_k\rangle$ and $\langle c\rangle$ means filling in the names of $e_k$ and $c$, respectively.

Besides, the information of a concept is also related to its sub- and super-concepts. Therefore, for each triple $(c_i, \textit{subclass of}, c_j)\in \mathcal{T}_c$, we also construct two queries with answers for $M_{sc}$: 
\begin{itemize}
    \item ``$\langle c_i\rangle$ || \textit{subclass of}'' $\rightarrow$ ``$\langle c_j\rangle$'';
    \item ``$\langle c_j\rangle$ || \textit{contains subclass}'' $\rightarrow$ ``$\langle c_i\rangle$''.
\end{itemize}

Finally, the information of a relation can be learned from its name and the elements connected by it. Therefore, for each $r\in R_l$, we sample $K$ triples $(e_i, r, e_j)$ from $\mathcal{T}_l$, choose $c_i$, $c_j$ such that $(e_i, \mathrm{instance\_of}, c_i), (e_j, \mathrm{instance\_of}, c_j)\in T_e$, and use each $(e_i, c_i, r, e_j, c_j)$ to construct three queries with answers: 
\begin{itemize}
    \item ``$\langle e_i\rangle$ | $\langle c_i\rangle$ || $\langle r\rangle$ | \textit{forward}'' $\rightarrow$ ``$\langle c_j\rangle$ | $\langle e_j\rangle$'';
    \item ``$\langle e_j\rangle$ | $\langle c_j\rangle$ || $\langle r\rangle$ | \textit{backward}'' $\rightarrow$ ``$\langle c_i\rangle$ | $\langle e_i\rangle$'';
    \item ``$\langle e_i\rangle$ | $\langle c_i\rangle$ || \textit{what relation} || $\langle c_j\rangle$ | $\langle e_j\rangle$'' $\rightarrow$ ``$\langle r\rangle$''.
\end{itemize}
We empirically find that including $c_i$, $c_j$ benefits the information encoding for both concepts and relations.

Let the set of all generated queries and answers be $D_{sc}=\{(q_i, a_i)\}_{i=1}^l$, then $m_{sc}$ is trained to minimize
\begin{equation}
    \mathcal{L}_{sc}=-\sum\limits_{(q_i, a_i)\in D_{sc}}\log P(a_i | q_i),
\end{equation}
where $P(a_i|q_i)$ is the likelihood of $M_{sc}$ generating $a_i$ given $q_i$, defined by token-level cross entropy. Note that the learning of $m_{sc}$ does not rely on any additional data except the KB itself, so we can train a schema plugin for any KB.

\subsubsection{Learning of PI Plugin}
As illustrated in Fig.~\ref{fig_framework}(c), to learn the PI plugin $m_{PI}$, we first train individual schema plugin $m_{sc}^{S_i}$ for each $\mathcal{KB}^{S_i}$. After that, given $(x_j^S, y_j^{S_1}, \dots, y_j^{S_N})\in D_a^S$, where $x_i^S$ is a question and $y_j^{S_i}$ is the golden program for $x_j^S$ on $\mathcal{KB}^{S_i}$, we train $m_{PI}$ by feeding $x_i^S$ to $M_{PI}^{S_1}, \dots, M_{PI}^{S_N}$ and requiring them to generate $y_j^{S_1}, \dots, y_j^{S_N}$, respectively. Here, $M_{PI}^{S_i}=\mathrm{Plug}(\mathrm{fz}({M}), \{\mathrm{fz}(m_{sc}^{S_i}), m_{PI}\})$. The overall objective can be formulated as:
\begin{equation}
    \mathcal{L}_{PI}=-\sum_{(x_j^S, y_j^{S_1}, \dots, y_j^{S_N})\in D_a^S}\sum_{i=1}^N \log P_i(y_j^{S_i} | x_j^S),
\end{equation}
where $P_i(y_j^{S_i} | x_j^S)$ is the likelihood of $M_{PI}^{S_i}$ generating $y_j^{S_i}$ given $x_j^S$, defined by token-level cross entropy. To generate programs conforming to different schemas given the same question, $m_{PI}$ must learn to (1) choose correct functions according to the compositional structure of the question; (2) extract and utilize question-relevant schema information for argument determination from the corresponding schema plugin, because it is the only difference among $M_{PI}^{S_1}, \dots, M_{PI}^{S_N}$.  

\subsubsection{Plugin Transfer}
Once the PI plugin $m_{PI}$ is trained, we directly transfer it to $\mathcal{KB}^T$ as in Fig~\ref{fig_framework} (d), and let $M_{PI}^T=\mathrm{plug}(M, \{m_{sc}^T, m_{PI}\})$ be the PI model for $\mathcal{KB}^T$. Here, $m_{sc}^T$ is the trained schema plugin for $\mathcal{KB}^T$ using the method in Sec.~\ref{sec_schema_plugin}.
Since $m_{sc}^T$ and $m_{sc}^{S_i}$ are trained with the same tasks, we expect that they encode schema information into their parameters in similar ways~\cite{intrinsic-dimension, prompt_transfer}, so $m_{PI}$ can also extract schema information from $m_{sc}^T$ to help PI over $\mathcal{KB}^T$.  Besides, to guarantee $M_{PI}^T$ generating valid programs which do not cause execution error or return an empty answer, we adopt constrained decoding, i.e., after $M_{PI}^T$ generates $f_1(arg_1), \dots, f_t(arg_t)$,  we enumerate all the valid $f_{t+1}(arg_{t+1})$ following the method of ~\citet{pangu} and restrict $M_{PI}^T$ to only generate one of them. More details are in Appendix~\ref{app_constrained_decoding}. We also use beam search to retain top-k programs during decoding to provide $M_{PI}^T$ with a more global view.
\section{Experiments}
\subsection{Datasets}
\noindent\textbf{Source Domain.} We use KQA Pro~\cite{kqapro} as the source domain datasets. It provides 117,970 questions with diverse compositional structures and corresponding programs based on a subset of Wikidata~\cite{wikidata}. 

\noindent\textbf{Target Domain. } We use WebQSP~\cite{webqsp}, GraphQ~\cite{graphq}, GrailQA~\cite{grailqa}, MetaQA~\cite{metaqa} and SoAyBench~\cite{soay} as the target domain datasets. Among them, WebQSP, GraphQ, and GrailQA are based on Freebase~\cite{freebase}. Their KBs contain a large number of schema items and can evaluate the effectiveness of KB-Plugin for large-scale KBs. MetaQA and SoAyBench are two datasets in movie and academic domains, respectively, and can evaluate the effectiveness for specific domains. For MetaQA, since most of the relations in its KB have been covered by KQA Pro, we remove these relations and relevant question-program pairs from KQA Pro to avoid data leakage. For SoAyBench which is originally a tool-using dataset based on Aminer~\cite{aminer} APIs, we construct its KB by collecting relevant data from these APIs. Table~\ref{tab_overlap} shows the statistics of these datasets and their overlap with source KBs generated from KQA Pro. Most schema items in the target KBs are unseen in source KBs and most test cases also involve unseen schema items. 
\begin{table}[t]
    \centering
    \scalebox{1.0}{
        \small
        \setlength{\tabcolsep}{3pt}
        \begin{tabular}{l|cc|cc|cc}
        \toprule
        \textbf{Dataset}          & $|\mathcal{R}|$ & $|\mathcal{R}_{u}|$ & $|\mathcal{C}|$ & $|\mathcal{C}_{u}|$ & $|\mathcal{D}^\text{test}|$  & $|\mathcal{D}_u^\text{test}|$ \\ \midrule
        KQA Pro    & 1209         & -            & 794         & -            & -     & -        \\
        WebQSP    & 412          & 296          & 446         & 363          & 1639  & 1083     \\
        GraphQ    & 9569         & 8931         & 7298        & 7004         & 2395  & 2340     \\
        GrailQA(dev)   & 3938         & 3524         & 2018        & 1868         & 6763  & 6578     \\
        GrailQA(test)   & 3938         & 3524         & 2018        & 1868         & 13231  & -     \\
        MetaQA    & 9            & 9            & 9           & 3            & 39093 & 39093    \\
        SoAyBench & 17           & 11             & 5           & 3            & 792   & 756     \\ \bottomrule
        \end{tabular}
    }
    \caption{Statistics for source and target domain datasets and their overlaps with 16 source KBs generated from KQA Pro. $|\mathcal{R}|$ / $|\mathcal{C}|$ denotes the number of relations / concepts in their KBs. $|\mathcal{R}_u|$ / $|\mathcal{C}_u|$ denotes the number of relations / concepts unseen in the source KBs. $|\mathcal{D}^\text{test}|$ and $|\mathcal{D}_u^\text{test}|$ denotes the numbers of test cases and test cases that involve unseen schema items, respectively. }
    \label{tab_overlap}
\end{table}

\subsection{Baselines}
For WebQSP, GraphQ, GrailQA, and MetaQA, we mainly compare KB-Plugin with low-resourced PI methods including (1) few-shot program generation methods \textbf{Pangu}~\cite{pangu} and \textbf{KB-BINDER}~\cite{kb-binder}; (2) few-shot data generation method \textbf{APS}~\cite{flexkbqa}; (3) program transfer method \textbf{ProgramTrans}~\cite{program_transfer}, where we adopt its results without fine-tuning on target KBs for fair comparison. In addition, we also provide the results of several representative supervised models for comparison. 

For SoAyBench, we choose tool-using methods that were evaluated on it as baselines, including \textbf{DFSDT}~\cite{toolllm} and \textbf{SoAy}~\cite{soay}. These methods solve questions by prompting LLMs to call Aminer APIs in specific orders via in-context learning. Their processes of determining the composition of APIs and filling in arguments for each API can also be viewed as program induction.

We provide detailed descriptions of all the baselines and our evaluation metrics in Appendix~\ref{app_baseline}.

\subsection{Implementation Details}
In experiments, we use Llama2-7B~\cite{llama2} as the backbone LLM of KB-Plugin and set the rank $r$ of LoRA to 16. The number of parameters of each plugin is consequently 40M, which is extremely lightweight. The number of generated source KBs is set to 16 to balance performance and training efficiency. The sampling number $K$ in schema plugin learning is set to be 500, 500, 50, 100, 3000, and 1000 for KQA Pro, WebQSP, GraphQ, GrailQA, MetaQA, and SoAyBench, respectively, to limit the size of the constructed data for schema plugin learning. We use beam size 5 for all experiments. More details can be found in Appendix~\ref{app_implementation}.
\begin{table}[t]
    \centering
    \small
    \setlength{\tabcolsep}{3.7pt}
    \begin{tabular}{l|cccc}
    \toprule
    \multirow{2}{*}{\textbf{Method}}      & \multirow{2}{*}{\textbf{WebQSP}}        & \multirow{2}{*}{\textbf{GraphQ}}  & \multicolumn{2}{c}{\textbf{GrailQA}} \\ 
     & & & Test & Dev \\\midrule
    \multicolumn{5}{l}{\textit{Supervised}}                                            \\ \midrule
    QGG                  & 74.0                  & -               & 36.7     &    -    \\
    BERT+Ranking         & -                     & 25.0            & 58.0     &    -    \\
    ArcaneQA             & 75.6                  & 31.8            & 73.7     &    76.8    \\
    RnG-KBQA             & 75.6                  & -               & 74.4     &    76.9   \\\midrule
    \multicolumn{5}{l}{\textit{Low-resourced}}                                        \\ \midrule
    ProgramTrans     & 53.8$^*$                 & -               & -      &    -      \\
    APS                  & 51.1                  & -               & 57.7 & 62.1           \\
    KB-BINDER & 53.2                  & 39.5            & 56.0        &  -   \\
    Pangu     & 54.5                  & 43.3            & \textbf{62.7}  &  -  \\
    \textbf{KB-Plugin}            & \textbf{57.2 / 61.1}$^*$ & \textbf{49.5}   & \textbf{62.7} & \textbf{65.0}  \\ 
    \textbf{\ \ w/o schema plugin} & 41.0           &  42.8        &   -      & 57.5   \\
    \textbf{\ \ w/} $m_{sc}^{S_0}$ & 48.0           &  37.9            &  -       & 51.0    \\
    
    \bottomrule
    \end{tabular}
\caption{F1 results on WebQSP, GraphQ, and GrailQA. $^*$ means using oracle topic entities. }
\label{tab_freebase}
\end{table}
\begin{table}[t]
\centering
\small
\begin{tabular}{l|ccc}
    \toprule
    \textbf{Method}    & \textbf{1-hop} & \textbf{2-hop} & \textbf{3-hop} \\ \midrule
    \multicolumn{4}{l}{\textit{Supervised}}                       \\ \midrule
    KV-Mem             & 96.2           & 82.7           & 48.9           \\
    PullNet            & 97.0           & 99.9           & 91.4           \\
    EmbedKGQA                & 97.5           & 98.8           & 94.8           \\
    TransferNet        & 97.5           & 100.0          & 100.0          \\ \midrule
    \multicolumn{4}{l}{\textit{Low-resourced}}                    \\ \midrule
    KB-BINDER          & 93.5           & 99.6           & 96.4           \\
    \textbf{KB-Plugin} & \textbf{97.1}  & \textbf{100.0} & \textbf{99.3} \\
    \textbf{\quad w/o schema plugin} &  92.6          &   99.0       &  98.9        \\
    \textbf{\quad w/} $m_{sc}^{S_0}$ &  90.4          &   93.6       &   88.6  \\
    \bottomrule
\end{tabular}
\caption{Hit@1 results on MetaQA.}
\label{tab_metaqa}
\end{table}
\begin{table}[t]
\centering
\small
\begin{tabular}{l|c}
\toprule
\textbf{Method}              & \textbf{Acc}  \\\midrule
DFSDT (gpt-3.5-turbo)    &        45.7       \\
DFSDT (gpt-4)            &        59.7       \\
SoAy (gpt-3.5-turbo)         &      67.7         \\
SoAy (gpt-4)                 &      88.7         \\
\textbf{KB-Plugin}                    & \textbf{90.8} \\
\textbf{\quad w/o schema plugin}            & 70.8          \\
\textbf{\quad w/} $m_{sc}^{S_0}$ & 64.0         \\ 
\bottomrule
\end{tabular}
\caption{Accuracy results on SoAyBench.}
\label{tab_soay}
\end{table}
\begin{table}[t]
\centering
\setlength{\tabcolsep}{4pt}
\small
\begin{tabular}{ll|cc}
\toprule
\textbf{Dataset}             & \textbf{Method} & $\mathcal{D}_\text{seen}^\text{test}$  & $\mathcal{D}_\text{unseen}^\text{test}$ \\ \midrule
\multirow{3}{*}{WebQSP}      & KB-Plugin                            & \textbf{64.9}        & \textbf{53.3}          \\
                             & w/o schema plugin                    & 47.6                 & 37.6                   \\
                             & \texttt{Gain}                                 & \textcolor[RGB]{76,187,23}{+17.4}                & \textcolor[RGB]{76,187,23}{+15.7}                  \\ \hline
\multirow{3}{*}{GraphQ}      & KB-Plugin                            & 40.0$^*$                & \textbf{49.7}          \\
                             & w/o schema plugin                    & \textbf{70.9}$^*$        & 42.2                   \\
                             & \texttt{Gain}                                 & \color{red}{-30.9$^*$}                & \textcolor[RGB]{76,187,23}{+7.5}                   \\ \hline
\multirow{3}{*}{GrailQA-dev} & KB-Plugin                            & \textbf{69.0}        & \textbf{64.8}          \\
                             & w/o schema plugin                    & 64.9                 & 57.3                   \\
                             & \texttt{Gain}                                 & \textcolor[RGB]{76,187,23}{+4.1}                 & \textcolor[RGB]{76,187,23}{+7.5}    \\
                             \bottomrule
\end{tabular}
\caption{F1 Results of KB-Plugin with and without schema plugin. $\mathcal{D}_\text{unseen}^\text{test}$ and $\mathcal{D}_\text{seen}^\text{test}$ denote the sets of test cases that involve and do not involve schema items unseen in the source KBs, respectively. $*$ means the results may not be indicative since there are only 55 cases in $\mathcal{D}_\text{seen}^\text{test}$ of GraphQ.}
\label{tab_unseen_schema}
\end{table}
\begin{table*}[t]
    \centering
\scalebox{0.7}{
    \begin{tabular}{cc}
    \toprule

\textbf{Question \uppercase\expandafter{\romannumeral1} } & \textbf{\it{Which airport to fly into Rome?}}  \\ \midrule
 Pangu & Find(Rome) Relate(\textcolor{red}{tourist attractions}) \,({\color{red}{\XSolidBrush}})
\\ 
KB-Plugin w/o schema plugin & Find(Rome) Relate(\textcolor{red}{country}) FilterConcept(\textcolor{red}{sovereign state}) \,({\color{red}{\XSolidBrush}})
\\ 
KB-Plugin & Find(Rome) Relate(\textcolor[RGB]{76,187,23}{transport terminus}) FilterConcept(\textcolor[RGB]{76,187,23}{airport}) 
 \, ({\color[RGB]{76,187,23}{\Checkmark}})\\
 \midrule
\multirow{2}{*}{Relevant Triples} & (London, \textcolor[RGB]{76,187,23}{transport terminus}, Luton airport), (London, instance of, citytown), \\
& (Luton airport, instance of, \textcolor[RGB]{76,187,23}{airport}) \\

\midrule

\textbf{Question \uppercase\expandafter{\romannumeral2} } & \textbf{\it{What role did Paul Mccartney play in the Beatles?}}  \\ \midrule
 Pangu &  \textcolor{red}{Find}(Paul Mccartney) \textcolor{red}{Relate}(instruments played)\,({\color{red}{\XSolidBrush}})
\\ 
KB-Plugin & \textcolor[RGB]{76,187,23}{Find}(Beatles) \textcolor[RGB]{76,187,23}{Relate}(member) \textcolor[RGB]{76,187,23}{Find}(Paul Mccartney) \textcolor[RGB]{76,187,23}{ReverseRelate}(member) \textcolor[RGB]{76,187,23}{And}() \textcolor[RGB]{76,187,23}{Relate}(role)
 \, ({\color[RGB]{76,187,23}{\Checkmark}})\\
 \midrule
\multirow{2}{*}{Source Domain Data Pair} & \textit{What is Jane Lynch's role in Glee?} \\
& \textcolor[RGB]{76,187,23}{Find}(Glee) \textcolor[RGB]{76,187,23}{Relate}(starring) \textcolor[RGB]{76,187,23}{Find}(Jane Lynch) \textcolor[RGB]{76,187,23}{ReverseRelate}(starring) \textcolor[RGB]{76,187,23}{And}() \textcolor[RGB]{76,187,23}{Relate}(character role)\\

\bottomrule
 
    \end{tabular}
    }
   \caption{Two typical questions from the test set of WebQSP that KB-Plugin succeeds while Pangu fails. The incorrect functions and arguments are marked as red, while the correct ones are marked as green.}
     \label{tab_case}
\end{table*}
\subsection{Main Results}
The results are presented in Table~\ref{tab_freebase},~\ref{tab_metaqa} and~\ref{tab_soay}. Compared with Pangu, the SoTA PI method for low-resourced KBs, KB-Plugin improves the F1 score by 2.7\% and 6.2\% on WebQSP and GraphQ, respectively, and achieves comparable performance on GrailQA, despite Pangu using 25$\times$ larger model (175B Codex) and 100 annotated examples from each dataset. Moreover, Pangu needs to call Codex hundreds of times for a question to score each candidate program, while our model selects the optimal program via beam search, which is significantly faster and less costly. Besides, since ProgramTrans, KB-BINDER, and Pangu all link questions to schema items according to their names only, the superiority of KB-Plugin also demonstrates the benefits of aggregating additional schema information from relevant triples via schema plugin learning. KB-Plugin even surpasses several supervised models on GraphQ and GrailQA, which demand training using thousands of annotated samples from target KBs, showing the effectiveness of transferring prior knowledge from rich-resourced KBs.

On MetaQA and SoAyBench, KB-Plugin outperforms all the low-resourced baselines even though they use more powerful LLMs (i.e., Codex, gpt-3.5-turbo, and gpt-4), indicating that our framework also performs well for domain-specific KBs. In particular, KB-Plugin achieves strong performance on par with supervised SoTAs on MetaQA even if it does not see any target relations from the source domain. 

\subsection{Ablation Study}
To demonstrate the effect of schema plugins, we remove them from our framework, i.e., we directly train a PI plugin using the source domain data and transfer it to the target KBs without training any schema plugins. According to Table~\ref{tab_freebase},~\ref{tab_metaqa}, ~\ref{tab_soay}, and~\ref{tab_unseen_schema}, the performance of KB-Plugin without schema plugins is severely degraded, especially on the test cases that involve schema items unseen in the source KBs. The experimental results illustrate that (1) direct PI transfer is difficult due to the substantial difference between the schemas of source and target KBs; (2) schema plugins of target KBs effectively encode adequate schema information via the triple completion task, and the PI plugin can extract and utilize question-relevant schema information from these schema plugins even though it is never trained with them. In addition, if we adopt the schema plugin of a source KB, e.g., $m_{sc}^{S_0}$, for the target KBs, the performance of KB-Plugin also drops heavily, showing the necessity of using matched schema plugin.

\begin{figure}[t]
    \centering
    \includegraphics[width=0.4\textwidth]{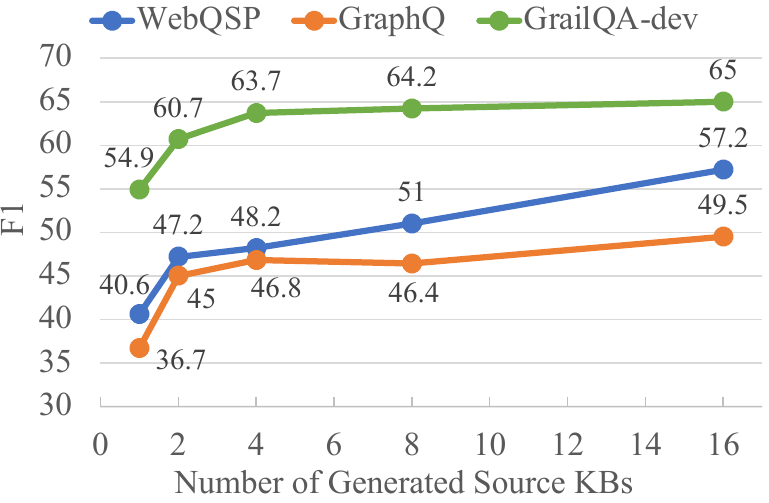}
    \caption{KB-Plugin performance with different numbers of generated source KBs.}
    \label{fig_num_kb}
\end{figure}

To show the rationality of our PI plugin learning method, we evaluate the performance of PI plugins trained with different numbers of generated source KBs on WebQSP, GraphQ, and GrailQA, and present the results in Fig.~\ref{fig_num_kb}. The PI plugin trained with only one source KB performs poorly, implying that it ignores the schema plugin entirely and directly learns PI over this source KB. Once there emerges a new source KB with a different schema, the performance of the trained PI plugin increases substantially, and there is an apparent trend that the performance will increase with more generated source KBs. These results prove that training the PI plugin over multiple source KBs succeeds in forcing the PI plugin to learn to extract and utilize schema information from different schema plugins, and the learned skill can be transferred to target KBs. 

\subsection{Case Study}
To better showcase the advantages of KB-Plugin over in-context learning PI methods, we present a case comparison between KB-Plugin and Pangu in Table~\ref{tab_case}. Question I shows the effect of schema plugin learning and utilization. Both Pangu and KB-Plugin without schema plugin struggle to predict the correct relation \textit{``transport terminus''} because it is unseen in the demo examples or source KBs. The complete KB-Plugin, however, effectively encodes the information that \textit{``transport terminus''} is a possible relation between \textit{``citytown''} and \textit{``airport''} into the schema plugin via completing relevant triples, and succeeds in predicting this relation by utilizing above information. Question II demonstrates the benefits of harnessing abundant program annotations from the source domain, where Pangu produces a program with incorrect function composition because none of its demo examples has a similar compositional structure, while KB-Plugin induces the correct program by utilizing prior knowledge learned from the source domain. Further analysis can be found in Appendix~\ref{app_unseen_structure} and~\ref{app_error_analysis}.

\section{Conclusion}
We propose KB-Plugin, a plug-and-play framework that enables LLMs to induce programs over any low-resourced KB by learning two types of pluggable modules: KB-specific schema plugin and KB-transferable PI plugin. KB-Plugin achieves better or comparable performance on five heterogeneous KBQA datasets with much smaller backbone LLMs compared to SoTA PI methods for low-resourced KBs, demonstrating its effectiveness for both large-scale and domain-specific KBs. 
Ablation study and case study also prove the rationality and further showcase the advantage of KB-plugin.

\section{Limitations}
We discuss several limitations of KB-Plugin in this section: (1) In the experiments, we only adopt Llama2-7B as our backbone model due to limited computing resources. Actually, KB-Plugin is model-agnostic and can also be applied to more language models with various sizes and architectures. (2) KB-Plugin requires that the source domain dataset covers questions with diverse various compositional structures, and performs relatively poorly for questions whose compositional structures are unseen in the source domain dataset though they are rare (see Appendix~\ref{app_unseen_structure} for details). Future research can focus on improving the transferability of KB-Plugin across compositional structures. In practice, we can also continue to train the PI plugin using some self-training methods such as EGST~\cite{flexkbqa} to adapt to these questions. (3) In this work, since both training and evaluation of KB-Plugin require annotated KBQA datasets, we can only take a single dataset KQA Pro as the source dataset and take other datasets as the target datasets, which may limit the upper bounds of KB-Plugin. 
In the realistic scenario where we need to apply KB-Plugin for a new KB, we can take all these KBQA datasets as the source domain datasets so that the trained source schema plugins would be more diverse and the trained PI plugin would also have stronger transferability and generalizability. 

\section{Ethical Considerations}
Though our framework (as well as other PI methods) can effectively reduce the probability of LLMs generating inaccurate answers when faced with questions involving uncommon knowledge, it may still make mistakes if the induced programs are incorrect. In addition, there is a risk of being hacked through targeted means such as injecting harmful or nonfactual knowledge into the KBs. Hence additional care and protective measures should be taken if our framework is deployed in user-facing applications. 

All the datasets and encyclopedias used in this work are publicly published with permissible licenses.

\bibliography{custom}

\newpage
\appendix
\section{Details of KoPL Functions}
\begin{table}[t]
\centering
\small
\scalebox{0.74}{
\begin{tabular}{ccc}
\toprule
\textbf{Function}   & \textbf{Input$\times$Args $\rightarrow$ Output} & \textbf{Description} \\\midrule
Find & $E\times \emptyset \rightarrow E$ & find an entity from the KB \\
FindAll & $\emptyset \times \emptyset \rightarrow E'$ & return all entities in the KB \\
Relate & $(E\cup E')\times R \rightarrow  E'$ & a single hop along a relation\\
ReverseRelate & $(E\cup E')\times R \rightarrow  E'$ & a reverse hop along a relation\\
FilterConcept & $E'\times C \rightarrow E'$ & return entities in a concept \\
And/Or & $(E', E')\times \emptyset \rightarrow E'$ & intersection/union of two sets \\
Argmax/Argmin & $E'\times R\rightarrow E'$ & superlative aggregations \\
LT/LE/GT/GE & $E\times R\rightarrow E'$ & $<$ / $\le$ / $>$ / $\ge$ \\
Count & $E'\times \emptyset \rightarrow N$ & set cardinality \\
\bottomrule
\end{tabular}
}
\caption{KoPL functions used in this work. $E$: entity; $E'$: a set of entities; $R$: relation; $C$: concept; $N$: integer.}
\label{tab_kopl}
\end{table}
We list KoPL functions used in this work in Table~\ref{tab_kopl}. We make some modifications to the original~\cite{kqapro} for conciseness. Except \texttt{Find} taking topic entities as the argument, other functions either have no arguments or take schema items (i.e., concepts or relations) as their arguments.

\section{Triple Sampling Strategy}
\label{app_sampling_strategy}
Let the given KB be $\mathcal{KB}=\{\mathcal{C}, \mathcal{E}, \mathcal{R}, \mathcal{T}\}$, where $\mathcal{T}=\mathcal{T}_e\cup\mathcal{T}_c\cup\mathcal{T}_l$. 
For each $e\in \mathcal{E}$, let $\operatorname{cnt}(e)$ be its popularity (i.e., the number of its occurrences in $\mathcal{KB}$).

When sampling \textit{``instance of''} triples for a concept $c\in \mathcal{C}$, we hope the sampled triples contain representative entities belonging to $c$, so we sort all $(e_k, \textit{instance of}, c)\in T_e$ in descending order of $\operatorname{cnt}(e_k)$ and select the first $K$ triples. 

When sampling relational triples for a relation $r\in \mathcal{R}_l$, we take both representativeness and diversity into account. Therefore, we sort all $(e_i, r, e_j)\in T_l$ in descending order of $\min(\operatorname{cnt}(e_i), \operatorname{cnt}(e_j))$ and select the first $K$ triples. 

\section{Details of Constrained Decoding}
\label{app_constrained_decoding}
In constrained decoding, after $M_{PI}^T$ generates $t$ function chunks $f_1(arg_1), \dots, f_t(arg_t)$,  we enumerate all admissible $f_{t+1}(arg_{t+1})$ as the candidate set $P_{t+1}$ following the definition of KoPL functions in Table~\ref{tab_kopl}, and constrain $M_{PI}^T$ to continue generating one of these candidate or generating the $\langle\mathtt{EOS}\rangle$ token to end the decoding process. 

Specifically, let $E_\text{topic}$ be the set of topic entities in the question obtained using off-the-shelf entity linkers~\footnote{Entity linking is not a major challenge for PI, and exhaustive fuzzy string matching~\cite{yao15} suffices to achieve a reasonable performance.}. At $t=0$, we enumerate \texttt{Find}($e$) for each $e\in E_\text{topic}$ as a candidate  in $P_1$. Specially, around 5\% of questions in GraphQ and GrailQA do not have a topic entity (e.g., “Who is the heaviest film director?" from GrailQA, whose target program is \texttt{FindAll}() \texttt{FilterConcept}(\textit{director})\texttt{SelectAmong}(\textit{weight kg}). For these questions, we follow Pangu~\cite{pangu} to start constrained decoding from \texttt{FindAll}()\texttt{FilterConcept}($c$), where $c$ is a topic concept provided by~\citet{arcaneqa}.

When $t>0$, we execute the current program $p_t=\langle f_1(arg_1), \dots, f_t(arg_t)\rangle$ to get its denotation (i.e., a set of entities) and also the concepts, forward relations, and backward relations that are reachable from the denotation. For each concept $c$, we enumerate \texttt{FilterConcept}($c$) as a candidate in $P_{t+1}$.  For each forward relation $r$, we enumerate \texttt{Relate}($r$) as a candidate. For each backward relation $r$, we enumerate \texttt{ReverseRelate}($r$) as a candidate, and also include \texttt{LT}($r$), \texttt{LE}($r$), \texttt{GT}($r$), and \texttt{GE}($r$) in $P_{t+1}$ if the denotation of $p_t$ is a numerical value such as a quantity or a date. In addition, candidates with superlatives can be enumerated as \texttt{Argmax}($r$) and \texttt{Argmin}($r$). Also, \texttt{Count}() can always be included to $P_{t+1}$. If there are multiple topic entities, we enumerate \texttt{Find}($e'$) as a candidate to add a new branch, where $e'\in E_\text{topic}$ is a topic entity not in $p_t$. When $p_t$ contains multiple branches, we enumerate \texttt{Or}() and \texttt{And}() as candidates to merge the last two branches.

\section{Experimental Setup}
\subsection{Details of Baselines and Evaluation Metrics}
\label{app_baseline}
The details of our baselines are as follows:

\noindent\textbf{Pangu}~\cite{pangu} utilizes potent LLM Codex~\cite{codex} to produce programs in a step-wise fashion via in-context learning. At each step, it first extends existing programs into new valid candidates by enumerating all possible next functions with arguments, then scores each candidate using Codex with several demonstrations and retains the top-k candidates.

\noindent\textbf{KB-BINDER}~\cite{kb-binder} first lets Codex generate several "draft" programs for a given question by imitating a few examples, then grounds the arguments in the drafts to the target KB using similarity search to produce hundreds of refined programs. The final answer is decided by the majority vote after executing all these refined programs.

\noindent\textbf{Automatic Program Sampling (APS)}~\cite{flexkbqa} utilizes gpt-3.5-turbo\footnote{https://platform.openai.com/docs/models/gpt-3-5} to translate automatically sampled programs based on a handful of templates into corresponding questions via in-context learning, and subsequently fine-tune a RnG-KBQA~\cite{rng_kbqa} PI model using the generated question-program pairs. 

\noindent\textbf{ProgramTrans}~\cite{program_transfer} is a program transfer method that first uses a seq2seq sketch parser to translate the question into a program sketch, then uses an argument parser to search suitable argument from the KB for each function. We adopt its results without fine-tuning on the target KBs for fair comparison.

\noindent\textbf{DFSDT}~\cite{toolllm} is the SoTA method for general tool using. To solve a question, it employs a LLM to call suitable tool APIs in depth-first order. At each step, the LLM can either (1) call the next API to proceed along a promising path or (2) undo the current call and call another API to expand a new path.

\noindent\textbf{SoAy}~\cite{soay} is the SoTA method on SoAyBench. Given a question, it employs a LLM to first select the most suitable plan (i.e., API combination) from a candidate pool, then write a Python program with branching and looping structure following the plan to call APIs to get the answer.

\noindent\textbf{Supervised Methods.} For WebQSP, GraphQ, GrailQA, and MetaQA, we also provide the fully supervised results of several representative models for comparison, including QGG~\cite{qgg}, BERT+Ranking~\cite{grailqa}, ArcnaeQA~\cite{arcaneqa}, RnG-KBQA~\cite{rng_kbqa}, KV-Mem\cite{kv-mem}, PullNet~\cite{pullnet}, EmbedKGQA~\cite{embkgqa} and TransferNet~\citet{transfernet}.

\noindent\textbf{Evalution Metrics.} Following these baselines, we use F1 for WebQSP, GraphQ, and GrailQA, use Hit@1 for MetaQA, and use Accuracy for SoAyBench.

\subsection{Implementation Details}
\label{app_implementation}
We train the schema plugins of the source and target KBs for 3 epochs and 1 epoch, respectively. The batch size and learning rate are set to be 128 and 1e-5, respectively. Besides, we train the PI plugin for 1 epoch with batch size 16 and learning rate 1e-5. For WebQSP, GraphQ, and GrailQA, we use the same off-the-shelf entity-linker as Pangu to find topic entities; For MetaQA, we follow our baselines to use oracle topic entities; For SoAyBench, we find topic entities using spaCy~\cite{spacy}.

\section{Analysis about Question Compositional Structures}
\label{app_unseen_structure}
\begin{table}[t]
\centering
\small
\begin{tabular}{l|ccc|ccc}
\toprule
\multirow{2}{*}{\textbf{Dataset}} & \multicolumn{3}{c|}{\textbf{Seen}}        & \multicolumn{3}{c}{\textbf{Unseen}}       \\
                                  & \textbf{Num} & \textbf{EM} & \textbf{F1} & \textbf{Num} & \textbf{EM} & \textbf{F1} \\ \midrule
GraphQ                            & 2148         & 71.0         & 52.8        & 247          & 15.4         & 20.4        \\
GrailQA                           & 6433         & 79.9         & 67.4        & 330          & 10.0         & 16.4       \\ \bottomrule
\end{tabular}
\caption{Performance of KB-Plugin on test cases whose compositional structures are seen and unseen in the source dataset KQA Pro. EM means the exact match of program sketch.}
\label{tab_unseen_structure}
\end{table}
For GraphQ and GrailQA, we translate their SPARQL programs to KoPL programs using GraphQ Trans~\cite{graphqir} and analyze the performance of KB-Plugin on the test cases whose question compositional structures (identified by program sketches) are seen and unseen in the source domain dataset KQA Pro, respectively. From the results in Table~\ref{tab_unseen_structure} we can see that (1) KQA Pro covers most of question compositional structures in the target dataset; (2) KB-Plugin correctly predicts the program sketches for over 70\% questions whose compositional structures are seen in the source domain dataset, implying that the mapping from questions to program sketches is largely independent of KB schemas and transferable across KBs, which is consistent with the findings of~\citet{program_transfer} and~\citet{kb-binder}; (3) KB-Plugin performs poorly on the questions with unseen compositional structures though they are relatively rare, indicating that more advanced transfer techniques across compositional structures remains to be explored. 

\section{Error Analysis}
\label{app_error_analysis}
We analyze 100 incorrect predictions (i.e., F1<1) randomly sampled from the dev set of GrailQA. The major errors are predicting wrong schema items (36\%). 
Specially, when facing several schema items with only subtle differences, e.g., \textit{``publisher''}(reverse) 
 v.s. \textit{``game version published''}, KB-plugin tends to prefer to choose the shorter one due to the inherent defects of beam search. 
Besides, 21\% errors are due to a wrong termination check where the model misses the last relation or predicts an additional function. There are also 5\% wrong function predictions. Apart from the above errors caused by our model, 27\% errors are caused by unidentified or wrongly identified topic entities during entity linking, 9\% errors are due to ambiguous or wrong annotations, and the remaining 2\% errors are due to the incompletion of KBs.

\end{document}